\documentclass[10pt,twocolumn,letterpaper]{article}

\usepackage[pagenumbers]{wacv} 

\definecolor{wacvblue}{rgb}{0.21,0.49,0.74}
\usepackage[pagebackref,breaklinks,colorlinks,allcolors=wacvblue]{hyperref}
\usepackage{amsmath}
\usepackage{amsfonts}
\usepackage{float}
\usepackage{graphicx}

\def\wacvPaperID{1894} 
\def\confName{WACV}
\def\confYear{2026}

\title{BeLLA: End-to-End Birds Eye View Large Language Assistant for Autonomous Driving}

\author{
Karthik Mohan$^{1}$,
Sonam Singh$^{2}$,
Amit Arvind Kale$^{2}$ \\
\small $^{1}$UC San Diego,
$^{2}$Robert Bosch Corporate Research India \\
\small \texttt{knmohan@ucsd.edu,
\{sonam.singh, amitarvind.kale\}@in.bosch.com}
}

\begin{document}
\maketitle
\begin{abstract}

The rapid development of Vision-Language models (VLMs) and Multimodal Language Models (MLLMs) in autonomous driving research has significantly reshaped the landscape by enabling richer scene understanding, context-aware reasoning, and more interpretable decision-making. However, a lot of existing work often relies on either single-view encoders that fail to exploit the spatial structure of multi-camera systems or operate on aggregated multi-view features, which lack a unified spatial representation, making it more challenging to reason about ego-centric directions, object relations, and the wider context. We thus present BeLLA, an end-to-end architecture that connects unified  $360^{\circ}$ BEV representations with a large language model for question answering in autonomous driving.  We primarily evaluate our work using two benchmarks - NuScenes-QA and DriveLM, where BeLLA consistently outperforms existing approaches on questions that require greater spatial reasoning, such as those involving relative object positioning and behavioral understanding of nearby objects, achieving up to +9.3\% absolute improvement in certain tasks. In other categories, BeLLA performs competitively, demonstrating the capability of handling a diverse range of questions.
\paragraph{}

\end{abstract}
\begin{figure*}[t]
    \centering
    \includegraphics[width=0.84\linewidth]{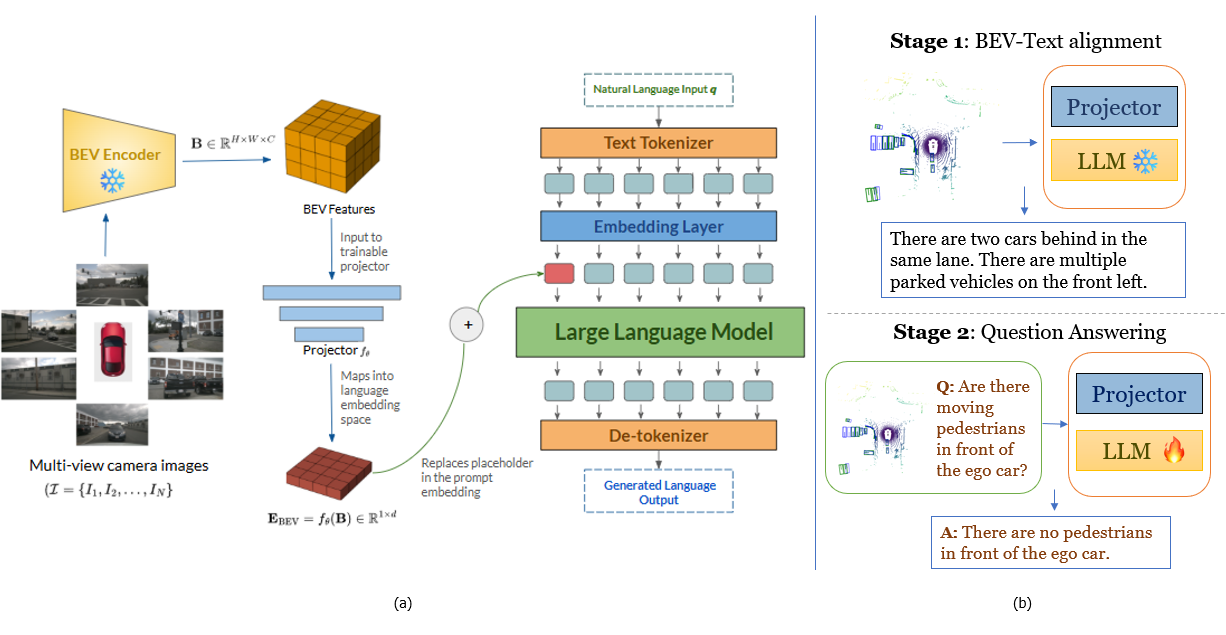}
    \caption{Overview of the BeLLA framework. A frozen BEV encoder processes multi-view camera images to obtain a unified BEV feature map \( \mathbf{B} \in \mathbb{R}^{H \times W \times C} \). Part (a) of the figure illustrates the general workflow shared across both pretraining and finetuning stages. During the pretraining phase, \( \mathbf{B} \) is passed through a trainable projector \( f_\theta \) to produce the projected embedding \( \mathbf{E}_{\text{BEV}} \in \mathbb{R}^{1 \times d} \), which is used for aligning the BEV text with its description using the frozen LLM. During the finetuning phase, a textual prompt \( \mathbf{\textit{q}} \)
 is embedded and fused with \( \mathbf{E}_{\text{BEV}} \) as input to the LLM, which is fine-tuned using parameter-efficient LoRA adapters for answer generation. While \( \mathbf{\textit{q}} \) corresponds to a natural language question during finetuning, it contains only a placeholder token in the pretraining stage, which is replaced by \( \mathbf{E}_{\text{BEV}} \).}
    \label{fig:overview}
\end{figure*}

\section{Introduction}

 Recent advances in multimodal foundation models have significantly expanded their utility in autonomous driving, shifting the field away from conventional rule-based systems toward more robust, data-driven architectures. These systems exhibit greater adaptability in complex, unseen scenarios. The integration of natural language capabilities further enables human-centric interaction, interpretability, and analysis of otherwise opaque black-box models.
 
Many recent approaches bridge visual inputs, such as images or video, with large language models (LLMs) to support reasoning and decision-making. These models are often fine-tuned for specific downstream tasks, such as visual question answering (VQA) \cite{antol2015vqa} or trajectory prediction from sensor data. For example, vision-language models (VLMs) \cite{liu2023visual,li2022blip} have been adapted to generate scene-level descriptions from prompt-based queries. DriveGPT4 \cite{xu2024drivegpt4} uses camera feeds to predict control signals (e.g., speed) and answer scene-specific questions. Similarly, \cite{yang2025lidar} integrates LiDAR data with an LLM to enable 3D captioning and spatial question answering.

In autonomous driving, accurately identifying the positions and potential trajectories of surrounding agents—such as vehicles and vulnerable road users (VRUs), including pedestrians—is essential for effective planning, prediction, and control. Achieving this requires a comprehensive understanding of the ego vehicle’s environment to ensure safe and efficient decision-making. Existing multimodal frameworks, including those proposed in \cite{xu2024drivegpt4,marcu2024lingoqa,gopalkrishnan2024multi}, commonly employ pretrained image encoders like CLIP \cite{radford2021learning} to extract features independently from each camera in a multi-camera setup, followed by feature concatenation before feeding into downstream modules. While these approaches benefit from the semantic richness of visual encoders like CLIP, they often fail to capture the spatial relationships and geometric structure inherent in multi-camera configurations.

In contrast, newer autonomous driving pipelines often rely on Bird’s Eye View (BEV) representations to produce a unified 360$^\circ$ scene abstraction, which serves as a standard input for downstream tasks in perception and planning. Motivated by this, we introduce BeLLA, an end-to-end approach that explicitly encodes $360^{\circ}$ scene understanding through a Bird’s Eye View (BEV) encoder and connects it with a large language model (LLM) for natural language interaction. The model takes as input multi-view images captured from vehicle-mounted cameras oriented in different directions (e.g., front and rear) and processes them using a frozen BEV encoder \cite{li2024bevformer}. We adopt the paradigm established in LLAVA \cite{liu2023visual}, incorporating a pretraining phase with a projection layer. Our experimental findings suggest that this initial alignment phase is critical for establishing meaningful correlations between BEV embeddings and natural language representations. We then fine-tune the model end-to-end for driving-oriented question answering (QA) tasks using datasets such as NuScenes-QA \cite{qian2024nuscenes} and DriveLM \cite{sima2024drivelm}, enabling question-answering grounded in perception, behavior, and planning. 
The primary contributions of this paper are as follows: 
\begin{itemize}
\item We propose BeLLA, an end-to-end multimodal architecture that bridges multi-camera $360^{\circ}$ input with language understanding through BEV-based features. This explicit encoding empowers the LLM to reason more effectively about scene actors and predict future outcomes.
\item We evaluate BeLLA on recent VQA datasets tailored for autonomous driving, encompassing tasks across perception, prediction, and planning. Our approach demonstrates significant improvements over strong baselines (up to $9.3\%$ absolute gain on certain tasks) while maintaining competitive performance on others, highlighting the efficacy of our method.
\item Through comprehensive ablation studies, we provide strong empirical evidence that aligning BEV encoder features with the LLM input is crucial, and demonstrate the importance of employing pretrained encoders tailored to the input modality.
\end{itemize}
\section{Related Works}

\subsection{Overview}

\subsubsection{MLLMs in Autonomous Driving}
Recent advancements in Multimodal Large Language Models (MLLMs) have demonstrated their potential in addressing key tasks in autonomous driving, such as motion planning, perception, and control, through the integration of diverse sensor modalities \cite{sha2023languagempc,mao2023gpt,wu2025language, xu2024vlm, zhou2024embodied, yang2023llm4drive}. GPT-Driver \cite{mao2023gpt} transforms vision inputs into structured language representations (e.g., bounding boxes, object classes) and utilizes the OpenAI GPT-3.5 model to directly predict driving trajectories. DriveGPT4 \cite{xu2024drivegpt4} employs a visual encoder to process single-view camera images, which are then interfaced with an LLM to generate control commands such as vehicle speed, along with handling driving-related question answering. LiDAR-LLM \cite{yang2025lidar} incorporates LiDAR point cloud data as input within the MLLM framework. Talk2BeV \cite{choudhary2024talk2bev} converts BEV representations into structured JSON formats suitable for language model consumption.

\subsubsection{VQA Task in Autonomous Driving}
The Visual Question Answering (VQA) task, introduced by \cite{antol2015vqa}, involves answering open-ended natural language questions based on the content of an input image. Accurate responses require the model to both comprehend the visual scene and perform appropriate reasoning. In the past few years, there have been several works exploring this in the context of autonomous driving.NuScenes-QA \cite{qian2024nuscenes} uses dedicated QA heads, such as MCAN and BUTD \cite{anderson2018bottom, yu2019deep}, to combine multi-modal features and directly predict answers in a classification setup. EM-VLM4AD~\cite{gopalkrishnan2024multi} utilizes multi-view inputs by concatenating camera-wise visual embeddings into a single representation, which is processed by a lightweight language model (T5) for question answering. Similarly, LingoQA~\cite{marcu2024lingoqa} introduces a driving-focused QA dataset along with an MLLM baseline based on a fine-tuned Vicuna model. More recently, the work by \cite{ding2024holistic} explores this task for video data, and introduces a new dataset benchmark - NuInstruct to tackle questions that require a greater understanding of temporal events.

\paragraph{Beyond Single Views: BEV in Driving Systems}
Prior work has predominantly relied on large pretrained vision encoders, which generate semantically rich embeddings owing to the vast and diverse datasets used during their training. Although such embeddings are generally effective for single-view understanding, they may fall short in scenarios requiring spatial reasoning about different directions around the ego vehicle, such as queries concerning the left or rear view. Addressing this limitation necessitates a representation that captures the complete spatial context of the scene. Our motivation for leveraging a Bird’s Eye View (BEV) representation stems from its widespread adoption in contemporary autonomous driving systems. BEV has emerged as a standard for encoding the vehicle’s surroundings and has demonstrated strong performance across core tasks such as segmentation and 3D object detection \cite{philion2020lift, ng2020bev, hu2021fiery, ma2024vision, li2024bevformer}, which are fundamental to downstream components such as planning and prediction. BEV representations have also been directly adopted as the primary input to end-to-end autonomous driving frameworks, which leverage the high-level BEV embeddings for unified policy learning, integrating perception, prediction, and planning in a fully differentiable end-to-end manner \cite{jiang2023vad, hu2023planning}.  

Building on these advances, recent efforts have started to explore linking BEV features with language models. BeLLA contributes to this direction with an end-to-end framework that directly projects raw BEV features into the LLM input space through a learnable embedding. The key novelty of our work lies in compressing the BEV representation into a single token and introducing a BEV–text pre-alignment stage, a combination that enables efficient training while yielding strong gains on spatial reasoning tasks.

\section{Method}

Our approach, as illustrated in Fig.\ref{fig:overview} follows a two-stage pipeline: (1) a pretraining phase that aligns Bird's Eye View (BEV) features with the corresponding textual descriptions via a projector module, and (2) an end-to-end finetuning stage where the language model is trained to answer driving-related queries grounded in BEV scene context.
\subsection{Pretraining for BEV-Language Alignment}
Let \( \mathcal{I} = \{I_1, I_2, \dots, I_N\} \) denote the set of synchronized multi-view camera images captured at a single timestep. These images are processed by a frozen BEV encoder, denoted as \( \text{BEVEnc}(\cdot) \), which produces a spatial feature tensor:
\begin{equation}
\mathbf{B} = \text{BEVEnc}(\mathcal{I}) \in \mathbb{R}^{H \times W \times C}
\end{equation}
where \( H \), \( W \), and \( C \) correspond to the height, width, and number of channels of the BEV feature map. We utilize BEVFormer \cite{li2024bevformer}, a transformer-based architecture that fuses multi-vew camera images into a unified BEV representation, as our BEV encoder. Our framework is agnostic to the choice of the BEV encoder, and other alternatives could also be used to potentially improve performance.

To transform \( \mathbf{B} \) into a token-compatible representation for a language model, we introduce a trainable projection module \( f_\theta \), parameterized by \( \theta \). This module generates a single-token embedding \( \mathbf{E}_{\text{BEV}} \in \mathbb{R}^{1 \times d} \), where \( d \) is the embedding dimension used by the language model:
\begin{equation}
\mathbf{E}_{\text{BEV}} = f_\theta(\mathbf{B})
\end{equation}
The projection module \( f_\theta \) is composed of a sequence of operations: a convolutional stack to extract hierarchical spatial features from \( \mathbf{B} \), followed by average pooling, flattening, and a multi-layer perceptron to produce a fixed-length embedding vector. Finally, the result is normalized and linearly projected into the language model’s token space. The complete transformation can be written as:

\begin{subequations}
\label{eq:projector_rewrite}
\begin{align}
\mathbf{X} &= \text{ConvStack}(\mathbf{B}) \\
\mathbf{z} &= \text{MLP}(\text{Flatten}(\text{AvgPool}(\mathbf{X}))) \\
\mathbf{E}_{\text{BEV}} &= \text{LayerNorm}(\text{Linear}(\mathbf{z}))
\end{align}
\end{subequations}

To train this alignment module, we construct a dataset of paired samples \( (\mathbf{B}, \mathbf{T}) \), where \( \mathbf{T} \) is a frame-level textual description. These descriptions are generated using NuScenes metadata and CAN bus signals, capturing object positions relative to the ego vehicle and their motion status. As BEV representations exhibit temporal redundancy across consecutive frames, we sub-sample every fourth frame during pretraining to reduce overlap in the training data. To further improve robustness and avoid overly repetitive phrasing, we also introduce lexical variation by sampling synonyms, ensuring that the textual descriptions exhibit natural diversity rather than sounding like near-duplicates.



The pretraining objective is to maximize the likelihood of generating the textual description \( \mathbf{T} \), conditioned on the projected BEV embedding \( \mathbf{E}_{\text{BEV}} \), using a frozen autoregressive language model:
\begin{equation}
\mathcal{L}_{\text{pretrain}} = - \sum_{t=1}^{T} \log P(y_t \mid y_{<t}, \mathbf{E}_{\text{BEV}})
\end{equation}
where \( \{y_t\}_{t=1}^T \) are the tokens comprising the description \( \mathbf{T} \). This alignment encourages the language model to associate BEV-derived spatial cues with semantic entities and their behaviors.

\subsection{Finetuning for Question Answering}

After the BEV-text projection module is pretrained, we fine-tune the model end-to-end for the downstream task of driving-oriented visual question answering (VQA). During this stage, the parameters of the BEV encoder remain frozen, while the projector \( f_\theta \) and the language model \( \mathcal{L}_\phi \), with parameters \( \phi \), are updated.

Given a scene at inference time, the set of multi-view camera inputs \( \mathcal{I} \) is passed through the BEV encoder to produce a feature map \( \mathbf{B} = \text{BEVEnc}(\mathcal{I}) \). This is projected via the learned projector \( f_\theta \) into a single-token embedding \( \mathbf{E}_{\text{BEV}} = f_\theta(\mathbf{B}) \in \mathbb{R}^{1 \times d} \), similar to the previous stage.

Let \( q \) denote a natural language question about the scene. This question is tokenized into a sequence of \( T \) embeddings \( \mathbf{Q} \in \mathbb{R}^{T \times d} \), where \( T \) is the number of tokens in the question. We reserve a fixed placeholder position in this sequence, which is replaced with the BEV embedding \( \mathbf{E}_{\text{BEV}} \). The resulting input sequence \( \mathbf{I}_q \in \mathbb{R}^{(T+1) \times d} \) is passed to the language model.

This architecture allows the language model to attend directly to the spatial information encoded in the BEV representation when generating answers. The model is trained to autoregressively decode the output answer tokens \( \{a_1, a_2, \dots, a_K\} \), using a cross-entropy objective defined as:

\begin{equation}
\mathcal{L}_{\text{FT}} = - \sum_{k=1}^{K} \log P(a_k \mid a_{<k}, \mathbf{I}_q)
\end{equation}

This finetuning process encourages the language model to produce responses that are not only linguistically fluent but also grounded in the physical and spatial dynamics of the driving scene. Empirically, we observe that injecting a BEV-derived token into the input stream improves the model's performance on questions requiring spatial reasoning, object interaction, and intent prediction. The prior pretraining stage facilitates more efficient learning in this phase by initializing the model with a strong spatial-language prior.

\section{Experiments}
\subsection{Datasets}
We evaluate our method on two publicly available question-answering datasets for autonomous driving: NuScenes-QA and DriveLM \cite{sima2024drivelm}. Both provide diverse QA pairs grounded in real-world driving scenarios and serve as comprehensive benchmarks for multimodal reasoning.
\subsubsection{NuScenes-QA}
NuScenes-QA is a large-scale VQA benchmark based on the NuScenes dataset, comprising over 460K QA pairs across 34K timesteps. Questions are categorized into five types:
\begin{itemize}
    \item \textbf{Exist} – Presence of specific objects, vehicles, or pedestrians in the scene
    \item \textbf{Count} – Quantitative counting of object instances
    \item \textbf{Object} – Identification based on spatial or semantic attributes
    \item \textbf{Status} – Queries on object condition or motion state
    \item \textbf{Comparison} – Relational reasoning between multiple objects
\end{itemize}

\subsubsection{DriveLM-NuScenes}
DriveLM-NuScenes includes 4072 training frames with over 377K QA pairs . The questions are categorized into the following categories: 
\begin{itemize}
    \item \textbf{Perception} – Location or attributes of key objects or the ego vehicle
    \item \textbf{Prediction} – Anticipated future states and reasoning about dynamics
    \item \textbf{Planning} – Foresight into the ego vehicle’s intended actions
    \item \textbf{Behavior} – Description of the ego vehicle’s current motion
\end{itemize}

\begin{table*}[ht]

\centering
\begin{tabular}{lcccccc}
\toprule
\textbf{Model} &  \textbf{Exist} & \textbf{Count} & \textbf{Object} & \textbf{Status} & \textbf{Comparison} & \textbf{Overall Accuracy $\uparrow$} \\
\midrule

CenterPoint + BUTD    & 84.1 & 21.3 & 49.2 & 55.9 & 69.2 & 58.1 \\
CenterPoint + MCAN    & 84.8 & 20.8 & 52.3 & 59.8 & \textbf{70.0} & 59.5 \\
\toprule
MSMDFusion + BUTD     & 85.1 & 23.2 & 52.3 & 59.5 & 68.5 & 59.8 \\
MSMDFusion + MCAN     & \textbf{85.4} & \textbf{22.2} & 54.3 & 60.6 & 69.7 & \textbf{60.4} \\
\toprule
BEVDet + BUTD         & 83.7 & 20.9 & 42.0 & 51.3 & 66.9 & 53.4 \\
BEVDet + MCAN         & 84.2 & 20.4 & 51.2 & 54.7 & 67.4 & 57.9 \\
BeLLA w/ LLama (ours)     & 80.9 & 20.5 & \textbf{54.8} & \textbf{69.9} & 62.7 & 59.6\\

\bottomrule
\end{tabular}
\caption{
Performance comparison across various approaches from \cite{qian2024nuscenes} on the six question types from the NuScenes-QA dataset. CenterPoint uses LiDAR-only inputs, MSMDFusion combines both LiDAR and camera modalities, while BEVDet and BeLLA rely solely on multi-camera inputs as the initial modality.
}

\label{table1}
\end{table*}

\begin{table*}[ht]
\centering
\begin{tabular}{lcccc}
\toprule
\textbf{Model} & \textbf{BLEU-4} $\uparrow$ & \textbf{METEOR} $\uparrow$ & \textbf{ROUGE-L} $\uparrow$ & \textbf{CIDEr} $\uparrow$ \\
\midrule
EM-VLM4AD Base \cite{gopalkrishnan2024multi} & 45.36 & 34.49 & 71.98 & 3.20 \\
EM-VLM4ADQ-Large \cite{gopalkrishnan2024multi} & 40.11 & 34.34 & 70.72 & 3.10 \\
DriveLM-Agent \cite{sima2024drivelm} & 53.09 & 36.19 & 66.79 & 2.79 \\
MiniDrive \cite{zhang2024minidrive}& 50.20 & 37.40 & 73.50 & 3.32 \\

BeLLA w/ Qwen (ours)  & 45.62 & 33.90 & 72.12 & 3.17 \\
\bottomrule
\end{tabular}
\caption{
Comparison of generative answer quality on the DriveLM test set using standard captioning and QA evaluation metrics. Vision-based baselines generally achieve higher scores due to their ability to leverage appearance cues such as color and texture. In contrast, BeLLA operates solely on BEV representations, yet still achieves fairly competitive performance. 
}

\label{table2}
\end{table*}
\begin{table*}[ht]
\centering
\begin{tabular}{lcccc}
\toprule
\textbf{Question Type} & \textbf{BLEU-4} & \textbf{METEOR} & \textbf{ROUGE-L} & \textbf{CIDEr} \\
\midrule
Perception & 32.92 & 28.34 & 71.36 & 2.95 \\
Planning   & 47.83 & 34.40 & 72.01 & 2.98 \\
Prediction & 35.24 & 34.43 & 78.45 & 3.33 \\
Behavior   & 79.70 & 53.21 & 84.34 & 4.01 \\
\bottomrule
\end{tabular}
\caption{Question-type-wise evaluation of the proposed approach on the DriveLM dataset.}
\label{table3}
\end{table*}

\subsection{Implementation details}
We conduct our experiments using three language models: LLaMA 3.2 (3B) \cite{touvron2023llama}, Phi 3.5 (3.8B) \cite{abdin2024phi}, and Qwen2 (7B) \cite{bai2023qwen}. Performance differences across datasets are analyzed in the ablation section, while the main results highlight the best-performing model for each benchmark. For the language models, we use LoRA-based parameter-efficient fine-tuning \cite{hu2022lora}, applying a rank of 64 and $\alpha$ of 128. The BEV projector and the language model are trained jointly using the AdamW optimizer \cite{loshchilov2017decoupled}, with learning rates of 1e-4 for the projector and 2e-4 for the language model, for 10 epochs. All experiments are conducted on 4 NVIDIA H100 GPUs with a batch size of 2. 

\subsection{Quantitative Results}
\subsubsection{NuScenes-QA results}
For the NuScenes-QA Dataset, we evaluate using Top-1 accuracy, consistent with the prior work. In Table \ref{table1}, we provide the results for our proposed approach alongside baselines, reporting accuracy for the entire dataset, as well as for individual question types. 

\par
Our method, using LLama 3.2B as the language model backbone, achieves an overall accuracy of 59.6\% across over 83k test samples. While this is not the highest overall score across all baselines, it is competitive with existing methods, many of which leverage richer sensor fusion setups combining LiDAR and camera data. 
In particular, BeLLA shows strong performance on the 'Status' type questions, which require understanding the motion state or behavior of the relevant scene elements. We attribute this to the effectiveness of our BEV-text pretraining, which exposed the model to rich status-oriented supervision, such as whether surrounding vehicles or pedestrians are moving or stationary. 
It is worth noting that our LLM-based approach does introduce more computational overhead than lightweight QA heads like MCAN or BUTD. However, we believe that the added interpretability and reasoning capacity of LLMs, such as multi-hop reasoning or justification of actions taken, justifies this trade-off in safety-critical domains like Autonomous Driving. 
\par
Comparison with other LLM variants are provided in the ablation section.

\subsubsection{Drive-LM results}

DriveLM represents a more generative QA task, featuring longer, free-form answers and a wider distribution of question types.

Compared to NuScenes-QA, DriveLM is less optimized for BEV-only reasoning because many questions depend on fine-grained visual cues such as object color or texture, which are not explicitly represented in BEV features. As a result, we expect lower performance on appearance-centric queries. Nevertheless, DriveLM also contains substantial categories—such as Planning and Prediction—that are highly spatial and temporally grounded.  In these areas, BeLLA remains much more effective. We evaluate model outputs using commonly adopted language generation metrics: BLEU-4, METEOR, ROUGE-L, and CIDEr, which are standard in vision-language and question answering benchmarks.\cite{papineni2002bleu,banerjee2005meteor, lin2004rouge, vedantam2015cider}

As shown in Table \ref{table2}, our model achieves BLEU-4 of 45.62, ROUGE-L of 72.12, METEOR of 33.9, and CIDEr of 3.17, which are competitive with existing baselines. Table \ref{table3} further breaks down performance by question type. 
As we had expected, 'Perception' questions are where BeLLA struggles the most, due to their dependence on fine-grained visual information, which is not preserved in BEV. On the other hand, BeLLA demonstrates strong results on 'Prediction' and 'Planning' based questions, as these rely much more on understanding scene dynamics, spatial layouts, and anticipated motion, all of which are well captured in the BEV representation. In addition, the high scores on 'Behavior' questions reflect the more structured, pattern-like nature of these answers, which the LLM learns to generate reliably based on BEV-encoded context. 
\begin{figure*}[htbp]
    \centering
    \includegraphics[width=0.8\linewidth]{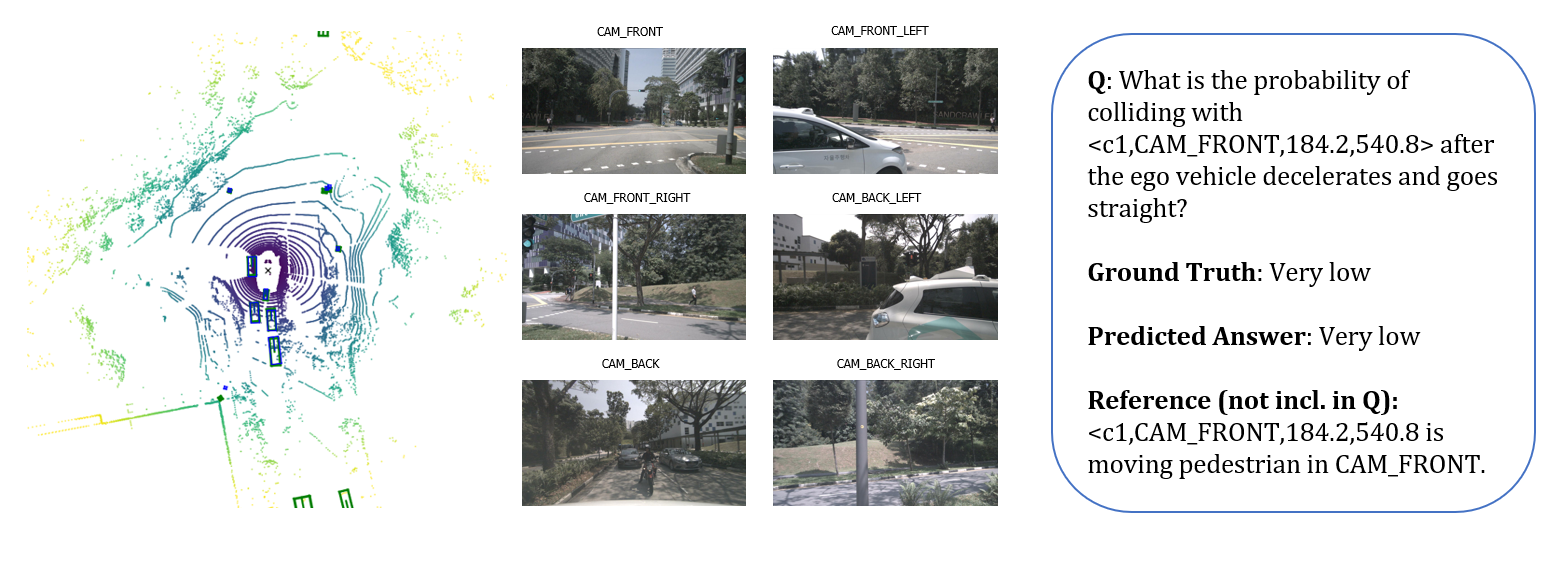}

    \includegraphics[width=0.8\linewidth]{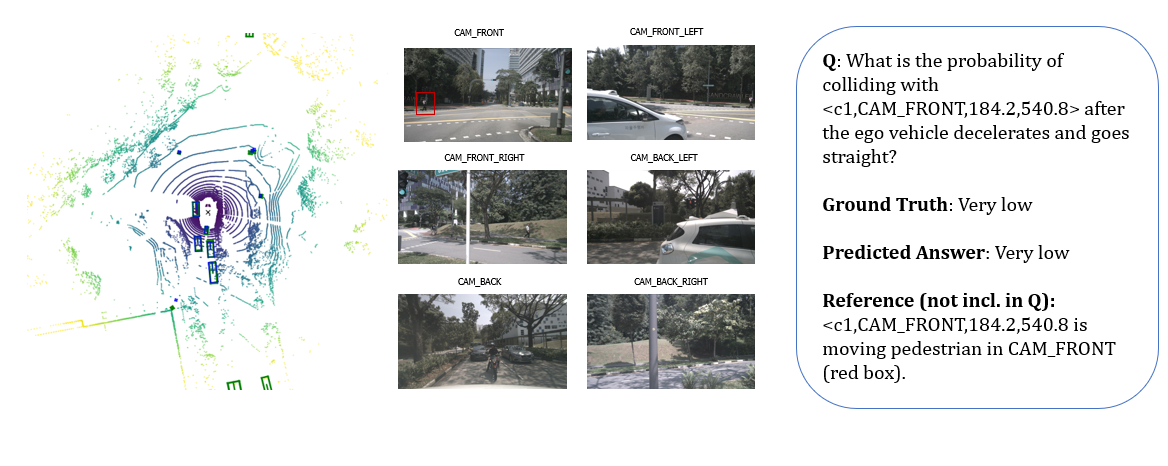}

    \includegraphics[width=0.8\linewidth]{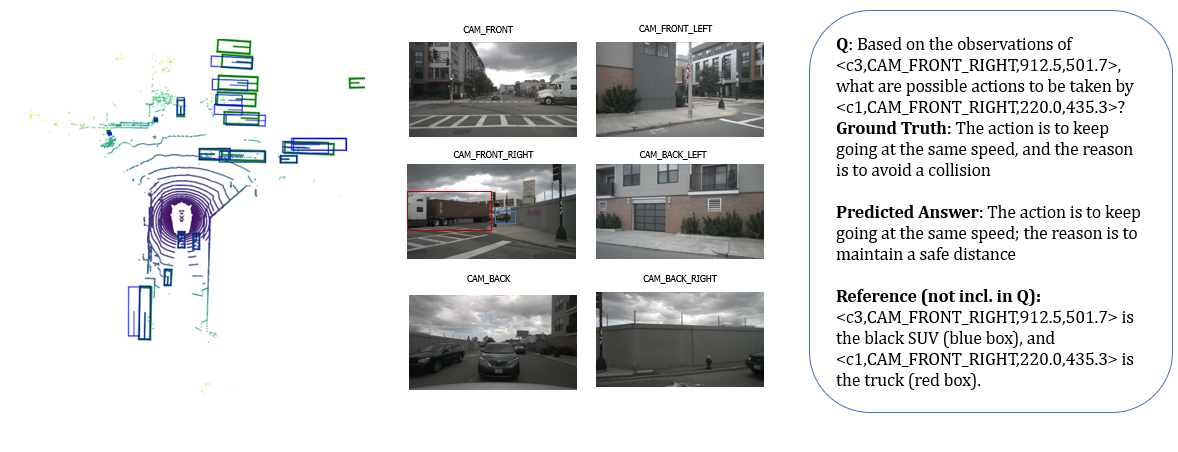}

  \includegraphics[width=0.8\linewidth]{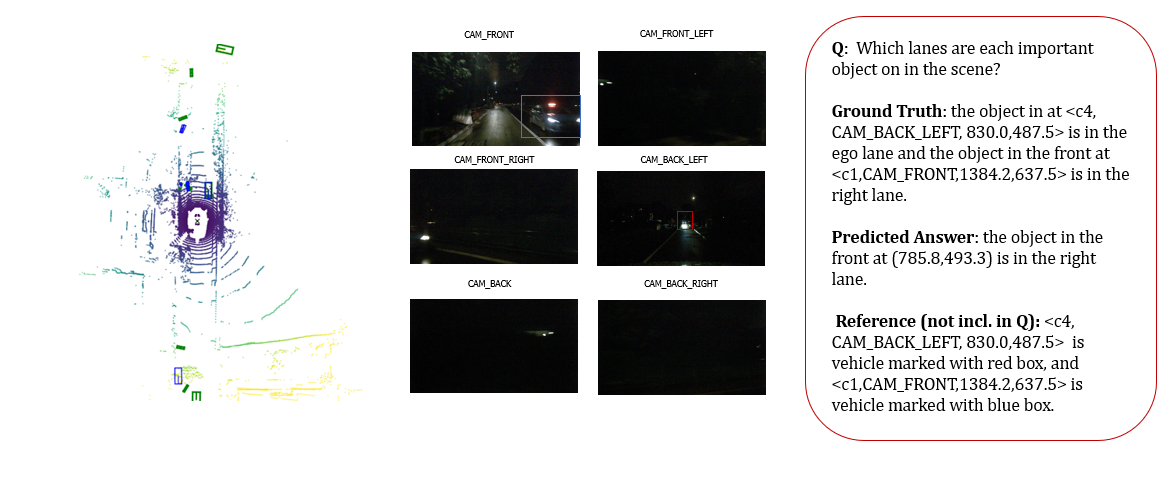}
    
    \caption{Qualitative results from the DriveLM benchmark: each example includes the rendered BEV map, the corresponding multi-view camera images, along with the natural language question, ground-truth answer, and the model's predicted answer. In some cases, the question refers to scene elements via spatial coordinates. To aid interpretation, we have manually mentioned those elements in the 'reference', as well as added a bounding box on the camera images. These cues are not provided to the model and are included solely for visualization purposes. Examples with a blue border denote correct (positive) responses, while the red-bordered example shows a failure case.}

    \label{fig:quali}
\end{figure*}
Overall, while BeLLA does not achieve the top score across every metric, its strengths are most evident in tasks requiring structured spatial reasoning and scene understanding - areas where the combination of BEV and LLMs provide an advantage.

\begin{table*}[ht]
\centering
\begin{tabular}{lcccccc}
\toprule
\textbf{Setting} & \textbf{Exist} & \textbf{Count} & \textbf{Object} & \textbf{Status} & \textbf{Comparison} & \textbf{Overall Acc. $\uparrow$} \\
\midrule
No Pretraining & 75.6 & 14.9 & 37.4 & 43.0 & 56.6 & 48.0 \\
Full Model& 80.9 & 20.5 & 54.8 & 69.9 & 62.7 & 59.6 \\
\bottomrule
\end{tabular}
\caption{Ablation on the effect of BEV-text pretraining using the NuScenes-QA dataset.}
\label{table4}
\end{table*}

\begin{table*}[!ht]
\centering
\begin{tabular}{lcccccc}
\toprule
\textbf{Projector} & \textbf{Exist} & \textbf{Count} & \textbf{Object} & \textbf{Status} & \textbf{Comparison} & \textbf{Overall Acc.} $\uparrow$ \\
\midrule
Linear & 75.1 & 14.1 & 37.8 & 50.3 & 56.3 & 49.4 \\
Conv.   & 76.7 & 14.2& 42.1 & 56.4 & 57.4 & 51.2 \\
\bottomrule
 Deeper Conv. & 80.9& 20.5& 54.8& 69.9& 62.7&59.6\\
\end{tabular}
\caption{Ablation on projector variants.}
\label{table5}
\end{table*}

\subsection{Ablation Studies}
To better understand the contribution of individual components within our framework, we conduct a number of ablation studies. Unless otherwise stated, all ablation experiments are performed on the NuScenes-QA dataset using the LLaMA 3.2 3B model. This setup allows us to isolate the impact of specific architectural or training choices while keeping the language backbone fixed. The only exception is the LLM backbone ablation, where we explicitly compare different LLM configurations.

\subsubsection{Effect of Pretraining}
To assess the importance of the pretraining stage, which focuses on aligning BEV representations with frame-level descriptions, we ablate this phase and directly finetune the LLama 3.2 3B model on the NuScenes-QA dataset. 
The results shown in Table \ref{table4} indicate a substantial performance drop when the pretraining stage is skipped. The most significant degradation occurs in the 'Status' category, which supports our earlier hypothesis that BEV-text pretraining is crucial for learning object-level motion and activity cues. These findings affirm the value of pretraining for enhancing downstream performance on temporally and semantically complex queries.
\subsubsection{Projector architecture}
The projector is a critical component of our framework, which transforms the BEV features into a token embedding that aligns with the LLM's input space. To understand the influence of this, we evaluate three variations of the projector module - a simple linear layer, a shallow convolutional model, and a deeper convolutional model. The latter serves as the main projector in our system and is used in all the primary experiments, and is described in detail in the Methods section.

The results shown in Table \ref{table5} can be directly compared with the main results obtained using the deeper convolutional model, which achieves an overall accuracy of 59.6. This gap between the lightweight models underscores the importance of having a deeper projector design for capturing the spatial and semantic richness of BEV Features. 
\subsubsection{LLM Backbone}
To assess the impact of different LLM backbones on performance, we evaluate three instruction-tuned models: LLaMA 3.2 (3B), Phi 3.5 (3.8B), and Qwen 2.5 (7B). On the NuScenes QA benchmark, which primarily involves single-word answers, LLaMA achieves the highest overall accuracy at 59.6, with Phi and Qwen slightly behind at 58.8 and 58.6, respectively.

For the DriveLM benchmark, which features longer and more free-form responses, Qwen (whose results are shown in Table \ref{table2}) performs best across most metrics, marginally outperforming LLaMA and Phi. While the gains are modest —typically within a single BLEU or ROUGE point— Qwen demonstrates stronger generative capabilities overall across the dataset.

These results suggest that while choice of LLM architecture and scale contribute modestly to performance, they are not the primary drivers of effectiveness in BEV-conditioned question answering. 
\subsection{Qualitative Analysis}
In Fig. \ref{fig:quali}, we present selected examples from the DriveLM dataset to qualitatively assess the responses generated by BeLLA. The model is capable of addressing a wide range of question types, including those related to collision avoidance, spatial understanding, and intent prediction. It demonstrates strong reasoning abilities in interpreting object dynamics, ego-relative positions, and planned behaviors using BEV features and multi-view context.
However, BeLLA struggles in scenarios where the BEV encoder fails to capture visual information - such as detecting objects in low-light or night conditions - as illustrated in the final example in Fig. \ref{fig:quali}. We also observe consistent limitations in answering questions that require precise spatial coordinate prediction, a challenge we found to be common among other LLM-based vision models as well. Additionally, the model underperforms on queries that depend on fine-grained visual attributes like traffic light color or vehicle appearance, which are not explicitly represented in BEV maps.

\section{Conclusion}
We introduce BeLLA, an end-to-end architecture that combines Bird's Eye View (BEV) representations with Large Language Models (LLMs) for question answering in autonomous driving. By leveraging a frozen BEV encoder and aligning its representations with language through a projector module, followed by the subsequent fine-tuning of the LLM, BeLLA enables effective spatial reasoning. Experimental results on NuScenes-QA and DriveLM demonstrate that our approach achieves competitive performance and excels in tasks requiring motion understanding and object interaction. The main limitations of our work are the difficulty in handling appearance-centric queries, due to BEV's lack of visual detail, and the lack of temporal modeling, as BeLLA currently operates on single-frame inputs. Future work could address the appearance limitations by incorporating complementary visual features from the raw camera views, as well as extend the model to video inputs for improved temporal reasoning and scene understanding.   
\bibliographystyle{plain} 
\bibliography{main}

\end{document}